\documentclass[11pt]{article}
\usepackage[preprint]{acl}
\usepackage{times}
\usepackage{latexsym}
\usepackage[T1]{fontenc}
\usepackage[utf8]{inputenc}
\usepackage{microtype}
\usepackage{inconsolata}
\usepackage{graphicx}
\usepackage{tabularx}
\usepackage{amsmath}
\usepackage{amsfonts}
\usepackage{algorithm}
\usepackage{algpseudocode}
\usepackage{booktabs} 
\usepackage{subcaption} 
\usepackage{caption}
\usepackage{float}
\title{Probing the Limits of Compressive Memory: \\ 
A Study of Infini-Attention in Small-Scale Pretraining}

\author{
  Ruizhe Huang, Kexuan Zhang, Yihao Fang\textsuperscript{\textdagger}, Baifeng Yu\textsuperscript{\textdagger} \\\\
  Huawei Technologies Canada Co., Ltd.
}

\begin{document}
\maketitle
{\renewcommand{\thefootnote}{\fnsymbol{footnote}}
\footnotetext[2]{Corresponding authors.}}
\begin{abstract}
This study investigates small-scale pretraining for Small Language Models (SLMs) to enable efficient use of limited data and compute, improve accessibility in low-resource settings and reduce costs. To enhance long-context extrapolation in compact models, we focus on Infini-attention, which builds a compressed memory from past segments while preserving local attention. In our work, we conduct an empirical study using 300M-parameter LLaMA models pretrained with Infini-attention. The model demonstrates training stability and outperforms the baseline in long-context retrieval. We identify the balance factor as a key part of the model performance, and we found that retrieval accuracy drops with repeated memory compressions over long sequences. Even so, Infini-attention still effectively compensates for the SLM's limited parameters. Particularly, despite performance degradation at a 16,384-token context, the Infini-attention model achieves up to 31\% higher accuracy than the baseline. Our findings suggest that achieving robust long-context capability in SLMs benefits from architectural memory like Infini-attention.
\end{abstract}

\section{Introduction}
Large Language Models (LLMs) deliver strong performance but suffer from the quadratic cost of attention \citep{vaswani2017attention}. Infini-attention \citep{munkhdalai2024infini} combines local attention with compressive memory for near-unbounded input lengths. Although scaling laws can be guiding principles for constructing larger and larger LLMs \citep{kaplan2020scaling}, attempts to purely scale are encountering cost-effectiveness and data availability limitations \citep{pmlr-v235-villalobos24a}. Consequently, there is renewed interest in Small Language Models (SLMs), which serve as a more flexible alternative to LLMs through their rapidly deployable computational efficiencies. SLMs are useful in constrained resource environments and agentic systems \citep{belcak2025small}, where a smaller model is needed for rapid inference and specialized deployment. This paradigm shift motivates our investigation into whether architectural innovations developed for large-scale models can be scaled down to enhance SLMs capabilities, even within the constraints of limited model size and context length.

We aim to scale down Infini-attention to adapt to a 300M-parameter LLaMA model, pretrained on the mostly short FineWeb dataset, to test whether memory-augmented attention yields benefits even when with small-scale pretraining.\footnote{The code is available in GitHub: \url{https://github.com/RRaAy-H/nanotron-infini}.}

Our contributions to the literature include:
\begin{itemize}
  \item We replace SLMs attention with Infini-attention, which accumulates cross-segment memory and balances with local attention, to study memory under short-sequence training.
  \item We analyze SLMs training dynamics, revealing larger loss fluctuations, gradient volatility, and early-layer memory concentration.
  \item We show that Infini-attention improves long-context extrapolation over the baseline, with supervised fine-tuning boosting performance.
\end{itemize}

\section{Related Work}
SLMs provide competitive accuracy to LLMs at much lower computational and deployment costs, making them suitable for agentic applications \citep{belcak2025small}, where lightweight models handle structured tasks while larger models reserved for complex reasoning. NVIDIA's Jet-Nemotron illustrates that architecture modifications, specifically freezing MLP layers and tuning attention, can be accomplished by post-pretraining and without depend solely on large-scale data \citep{gu2025jetnemotron}.  Similarly, Nemotron-Nano-9B-v2 applies FP8 training, multimodal data, and compression to allows single-GPU inference on the 128k-token context \citep{Basant2025NVIDIANN}, demonstrates the role of efficient KV-cache design.

Enhancing long-context reasoning is also essential for SLMs. Infini-attention that acquires memory across segments in a recurrent fashion and maintains local attention to extrapolate inputs with long sequence of tokens \citep{munkhdalai2024infini}. Taken together these developments, SLMs are positioned as a distinct paradigm emphasizing throughput, cost-effective, and low latency.

\section{Methodology}
\subsection{Baseline LLaMA-300M Architecture}
Our experiments are based on a 300M-parameter variant of LLaMA implemented in the Nanotron framework\footnote{\url{https://github.com/huggingface/nanotron}}. The base model has 12 decoder layers with hidden size of 1,024 and feed-forward dimension of 4,096. It uses 8 attention heads and key-value heads, with a maximum context length of 8,192 tokens. The vocabulary size is 49,152, with untied embeddings. The tokenizer used for both pre-training and fine-tuning is \texttt{lvwerra/the-tokenizer-v1}. We apply RMSNorm with $\epsilon=1\times 10^{-5}$, SiLU activation, and train with \texttt{bfloat16} precision.

\subsection{Infini-attention Integration}
We replace the standard causal self-attention module with Infini-attention \citep{munkhdalai2024infini}, with three key modifications. First, sequences exceeding 1,024 tokens are segmented, with memory accumulating information across segments. Second, past key–value pairs are compressed using $\mathrm{ELU}(x)+1$ for stability, where queries retrieve normalized weighted sums of stored values. Third, each attention head employs a trainable balance factor $\alpha \in [0,1]$ (via hard sigmoid) that interpolates between memory and local attention outputs:
\[
A = \alpha \cdot A_{\text{mem}} + (1-\alpha) \cdot A_{\text{local}}.
\]

The forward pass of Infini-attention is summarized in Algorithm~\ref{alg:infini}.  

\begin{algorithm}[hbt!]
\caption{Infini-attention Forward Pass}\label{alg:infini}
\begin{algorithmic}[1]
\small
\Require Input sequence $X$, segment length $S$, memory $M$, normalization $N$
\Ensure Output sequence $Y$
\State Split $X$ into segments $\{X_1, X_2, \dots, X_n\}$ of length $S$
\For{each segment $X_i$}
    \State Compute projections $Q, K, V \gets \text{Proj}(X_i)$
    \State Compute local attention $A_{\text{local}} \gets \text{Attention}(Q,K,V)$
    \If{memory is enabled}
        \State Retrieve $A_{\text{mem}} \gets \text{Retrieve}(Q, M, N)$
        \State Combine $A \gets \alpha \cdot A_{\text{mem}} + (1-\alpha) \cdot A_{\text{local}}$
        \State Update $(M,N) \gets \text{Update}(M,N,K,V)$
    \Else
        \State $A \gets A_{\text{local}}$
    \EndIf
    \State Project back $Y_i \gets W_o A$
\EndFor
\State Concatenate outputs $Y \gets [Y_1; Y_2; \dots; Y_n]$
\end{algorithmic}
\end{algorithm}

\subsection{Pre-training Dataset}
We pre-train on the \textbf{FineWeb sample-10BT} dataset \citep{NEURIPS2024_370df50c} for 30,000 steps. The corpus contains 14.9M documents, tokenized with \texttt{lvwerra/the-tokenizer-v1}, with a median length of 418 tokens, a mean of 716 tokens, and a standard deviation of 1,428 tokens. We set the training sequence length at 8,192 tokens with segment sizes of 1,024 tokens. Only 0.4\% of documents exceed this limit, so nearly all training sequences fall within just one segment.

\subsection{Pre-training Setup}
Both of models were pre-trained on the FineWeb sample-10BT dataset \citep{NEURIPS2024_370df50c}. Each sequence is truncated or padded to 8,192 tokens, with a global batch size of 4, resulting in 32,768 tokens per step and a total of $\sim$\textbf{983M} training tokens for 30,000 steps. We use AdamW optimizer with $\beta_1 = 0.9$, $\beta_2 = 0.95$, $\epsilon = 1\text{e}{-8}$, gradient clipping at 1.0, and weight decay 0.1. The learning rate is $6\times 10^{-5}$ with 500 warmup steps, followed by cosine decay to a floor of $6\times 10^{-6}$. 

\subsection{Baseline Adjustments}
The baseline architecture is identical to that of Infini-attention, but with memory disabled. Default learning rates caused vanishing gradients, and stronger clipping also failed. Raising the rate to $1.2\times10^{-4}$ stabilized training, suggesting Infini-attention implicitly regularize early gradient flow.

\section{Pre-training Results}
\label{sec:pre-training}
\subsection{Training Loss}
\begin{figure}[hbt!]
    \centering
    \includegraphics[width=\linewidth]{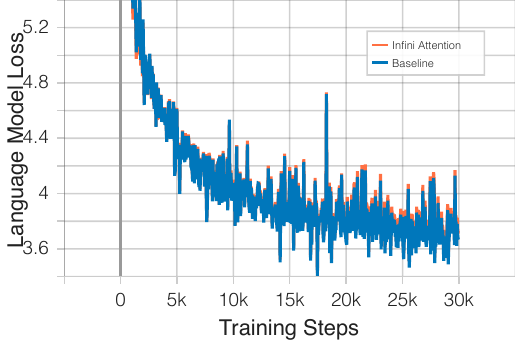}
    \caption{Training loss comparison between baseline and Infini-attention models.}
    \label{fig:loss}
\end{figure}

Figure~\ref{fig:loss} plots the training loss of 30,000 steps for both models. Both converge to a similar final loss, around 3.72 for Infini-attention and 3.68 for the baseline, which indicates comparable asymptotic performance. Throughout training, both curves has relatively large fluctuations, where neither model consistently outperforms the other in overall loss trajectory. The baseline occasionally dips slightly lower, whereas Infini-attention remains in a similar range, showing that both mechanisms yield comparable learning behavior under our setup.

\subsection{Gradient Norms}

\begin{figure}[hbt!]
    \centering
    \includegraphics[width=\linewidth]{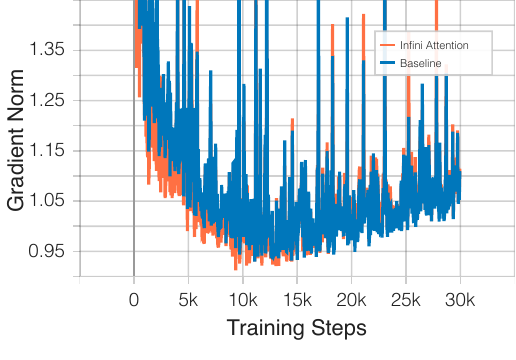}
    \caption{Gradient norm comparison between baseline and Infini-attention.}
    \label{fig:grad_norm}
\end{figure}

Figure~\ref{fig:grad_norm} illustrates that both models' gradient norms lie in a similar range overall. Infini-attention has slightly lower and more stable norms in the early training stages, whereas the baseline displays greater volatility. Particularly around the 10,000 step, the baseline experiences more spikes. Despite these localized differences, both models reflected relatively stable gradient dynamics over training.

\subsection{Balance Factors}
Infini-attention uses balance factors to interpolate between memory retrieval and local attention. On average, heads placed roughly roughly 30\% weight on memory and 70\% on local attention. This skew toward local attention aligns with the short-sequence bias of the FineWeb dataset. Additionally, memory use is concentrated in early layers, while deeper layers rely almost entirely on local attention. Full analysis is provided in Appendix~\ref{sec:appendix_memory}.

\section{Supervised Fine-Tuning}

\subsection{Fine-Tuning Dataset}
The fine-tuning data is from needle in a hay stack finetune dataset\footnote{\url{https://huggingface.co/datasets/nanotron/needle_in_a_hay_stack_finetune_dataset}}. This synthetic dataset tests long-context retrieval by placing a "needle" at various depths within a passage. The needle's position is set at 10\% increments of the total context length, which varies across 1,024 tokens, 4,096 tokens, 8,192 tokens, 16,384 tokens, and 32,768 tokens.

\subsection{Fine-Tuning Configuration}
The fine-tuning runs for 500 steps with a global batch size of 64. The learning rate was set to $7.5 \times 10^{-5}$, with a linear warmup over the first 50 steps followed by a cosine decay to a minimum rate of $3 \times 10^{-6}$. To maintain numerical stability, gradient accumulation was performed in FP32.

\section{Evaluations}
\label{sec:evaluations}

\subsection{Passkey Retrieval Evaluation}
\label{sec:eval}

We evaluate long-context capability for both models on the needle-in-a-haystack passkey retrieval dataset\footnote{\url{https://huggingface.co/datasets/nanotron/simple_needle_in_a_hay_stack}}, which is independent from the fine-tuning dataset.  
Table~\ref{tab:passkey_combined} compares zero-shot and fine-tuned performance across context lengths and depths.

\begin{table*}[htbp]
\centering
\resizebox{\textwidth}{!}{%
\begin{tabular}{lcccccc}
\toprule
& \multicolumn{6}{c}{\textbf{Context Length}} \\
\cmidrule(lr){2-7}
\textbf{Model} & \textbf{1,024}& \textbf{2,048}& \textbf{4,096}& \textbf{8,192}& \textbf{16,384}& \textbf{32,768}\\
\midrule
\multicolumn{7}{l}{\textit{Zero-shot (No Fine-tuning)}} \\
\midrule
Infini-attention & 100/100/100/100/100 & 100/100/100/100/100 & 1/0/0/3/0 & 0/0/0/0/0 & 0/0/0/0/0 & 0/0/0/0/1 \\
Baseline & 100/100/100/100/100 & 95/98/99/99/100 & 0/0/0/0/1 & 0/0/0/0/0 & 0/0/0/0/0 & 0/0/0/0/0 \\
\midrule
\multicolumn{7}{l}{\textit{After Fine-tuning (500 steps)}} \\
\midrule
Infini-attention (FT) & 96/93/95/100/99 & 88/98/99/98/99 & 61/96/99/69/1 & 29/90/0/0/0 & 45/0/0/0/0 & 48/0/0/0/1 \\
Baseline (FT) & 100/100/100/100/100 & 95/99/99/99/100 & 35/59/74/57/1 & 25/74/0/0/0 & 14/0/0/0/0 & 28/0/0/0/0 \\
\bottomrule
\end{tabular}%
}
\caption{Passkey retrieval zero-shot accuracy (\%) before and after fine-tuning. Each cell reports depth-wise accuracy at 0/25/50/75/100 percent of depths.}
\label{tab:passkey_combined}
\end{table*}

\paragraph{Zero-shot performance}
Without fine-tuning, both models retrieve perfectly at short contexts (less than 2,048 tokens), but performance drops beyond 4,096 tokens. Infini-attention shows a faint signal at 4,096 tokens and one success at 32,768 tokens, while the baseline fails, confirming short-sequence training limits long-range retrieval.  

\paragraph{After Fine-tuning}
With 500 steps of supervised fine-tuning, both models improve dramatically up to 4,096 tokens. Infini-attention was better at this length, maintaining high accuracy at intermediate depths. However, the gains do not hold much further due to repeated memory compressions over long sequences: at 8,192 tokens, retrieval succeeds only near the start of the sequence, and beyond 16,384 tokens both models fail almost entirely except needles at the start of the sequence. Even so, Infini-attention still outperforms the baseline in long-context retrieval.

\subsection{Standard Benchmark Evaluation}
To provide a broader assessment of model capabilities, we evaluated both the pretrained and fine-tuned models of the Infini-attention and baseline on a set of benchmarks.

\paragraph{General Reasoning Benchmarks}
As shown in Table~\ref{tab:general_benchmarks}, the Infini-attention model demonstrates better reasoning capabilities than the baseline model. The pretrained version achieves the highest accuracy on ARC-Challenge \citep{Clark2018ThinkYH} and WinoGrande \citep{10.1145/3474381}. After fine-tuning, it also had the top score on HellaSwag \citep{zellers2019hellaswag}. Additional results on the GLUE benchmark and the long-context Scrolls benchmark are available in Appendix~\ref{sec:appendix_benchmarks}.

\begin{table}[htbp]
\centering
\resizebox{\columnwidth}{!}{%
\begin{tabular}{lccc}
\toprule
\textbf{Model} & \textbf{ARC-Challenge} & \textbf{HellaSwag} & \textbf{WinoGrande} \\
\midrule
Baseline & 0.194 & 0.172 & 0.496 \\
& (0.012) & (0.039) & (0.044) \\
\midrule
Baseline (FT) & 0.204 & 0.183 & 0.496 \\
& (0.012) & (0.040) & (0.044) \\
\midrule
Infini-Attn & \textbf{0.251} & 0.172 & \textbf{0.602} \\
& (0.013) & (0.039) & (0.043) \\
\midrule
Infini-Attn (FT) & 0.241 & \textbf{0.204} & 0.526 \\
& (0.013) & (0.042) & (0.043) \\
\bottomrule
\end{tabular}%
}
\caption{Performance on general reasoning benchmarks. We report accuracy with standard errors in parentheses. Best results per task are highlighted in bold.}
\label{tab:general_benchmarks}
\end{table}

\section{Findings}
\label{sec:findings}
\paragraph{Dataset bias limits memory learning}
The median sequence of the FineWeb dataset was 418 tokens, which is far shorter than our segment length of 1,024. 
As a result, most training examples fit within a single segment, and the memory mechanism was rarely exercised.  
This likely explains the limited improvement in long-context retrieval, as the model had insufficient opportunities to learn cross-segment memory storage and retrieval.

\paragraph{Memory is concentrated in early layers}
Infini-attention heads in the first three layers strongly favored memory, while deeper layers relied almost entirely on local attention.  
This suggests long-context integration was handled early, with later layers focusing on local retrieval.  

\paragraph{Training stability benefits}
Infini-attention had a smoother convergence than the baseline and avoided vanishing gradients under the default learning rate.  
In contrast, the baseline model had to double its learning rate to achieve stable training.
This suggests that Infini-attention may serve as an implicit regularizer that improved the gradient flow.

\section{Conclusion and Future Work}
The research incorporated Infini-attention into a 300M-LLaMA model. The study demonstrated that Infini-attention can be trained effectively at smaller scales and addresses the vanishing gradient problem in the baseline model. The analyses indicated improved stability of optimization and memory specialization that was interpretable; however, the benefit for a long-context reasoning use case was less pronounced, due to the very short training sequence. Further research should use longer document datasets for training, up-scale models, and multi-task fine-tuning to assess how the memory could be used for long-context applications.

\section{Limitations}

Our work has several limitations that should be acknowledged.  
First, the training dataset (FineWeb sample-10BT) consists primarily of short text segments. As a result, the Infini-attention mechanism rarely encountered examples requiring cross-segment retrieval, limiting its ability to learn effective long-context memory usage.  
Secondly, the implementation of our experiments was performed on a relatively small model (300M parameters). This scale will be much smaller than state-of-the-art LLMs, and the absolute performance of both baseline and Infini-attention models was limited.  
Generated text often contained repetitions and incoherence, making it difficult to assess downstream quality beyond controlled retrieval tasks.  
Finally, supervised fine-tuning experiments were only using needle-in-a-haystack finetune dataset, which may not be diverse enough.
These limitations highlight the importance of training and fine-tuning dataset diversity, model scale, when developing and testing memory-augmented architectures like Infini-attention.

\bibliography{custom}

\appendix

\section{Appendix}
\label{sec:appendix}
\subsection{Balance Factors Analysis}
\label{sec:appendix_memory}
The balance factor plays a crucial role in determining the relative contribution of local attention and memory retrieval in Infini-attention. 

\subsubsection{Mean activated balance factor}
\label{sec:balance_mean}
\begin{figure}[hbt!]
    \centering
    \includegraphics[width=0.8\linewidth]{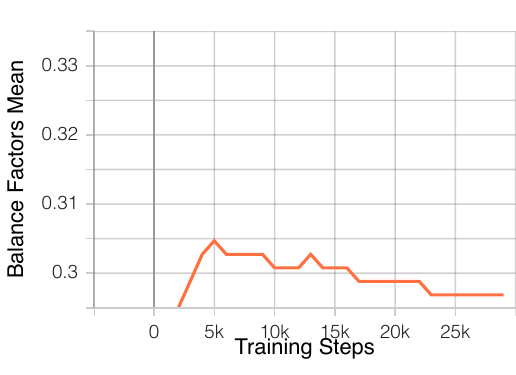}
    \caption{Mean activated balance factor across attention heads during Infini-attention training. The values converge around 0.30, reflecting a dataset bias toward shorter sequences where local attention is more heavily weighted than memory retrieval.}
    \label{fig:balance_mean}
\end{figure}

As shown in Figure~\ref{fig:balance_mean}, the mean value converges quickly to a stable value.
The convergence of balance factors around 0.30 indicates that the model appears to learn to prefer local attention compared with memory retrieval given the training distribution.

\subsubsection{Balance Factor Distribution}
\label{sec:balance_distribution}

\begin{figure}[hbt!]
    \centering
    \includegraphics[width=0.8\linewidth]{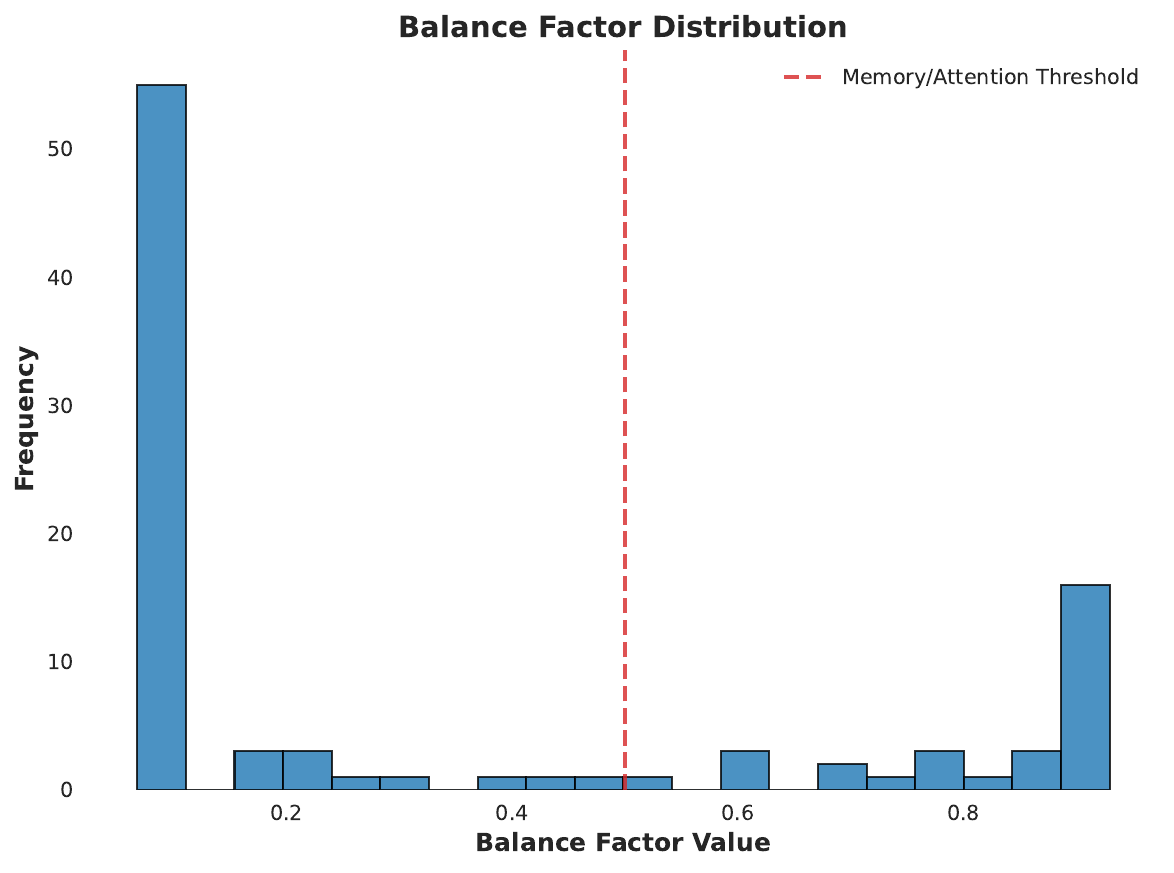}
    \caption{Distribution of balance factors across attention heads. The distribution shows significant variation, with most heads clustering around lower values (favoring local attention) while some heads specialize in memory retrieval with higher balance factors.}
    \label{fig:bf_distribution}
\end{figure}

The distribution of balance factors across attention heads provides insights into the heterogeneity of attention patterns within the model. Figure~\ref{fig:bf_distribution} shows this distribution, revealing the diversity in how individual heads learn to balance local and memory-based attention. The heavily skewed distribution toward lower balance factor values confirms that most attention heads prioritize local context, with only a subset specializing in long-range memory integration.

\subsubsection{Layer-wise Memory Preferences}
\label{sec:layer_preferences}

\begin{figure}[hbt!]
    \centering
    \includegraphics[width=0.8\linewidth]{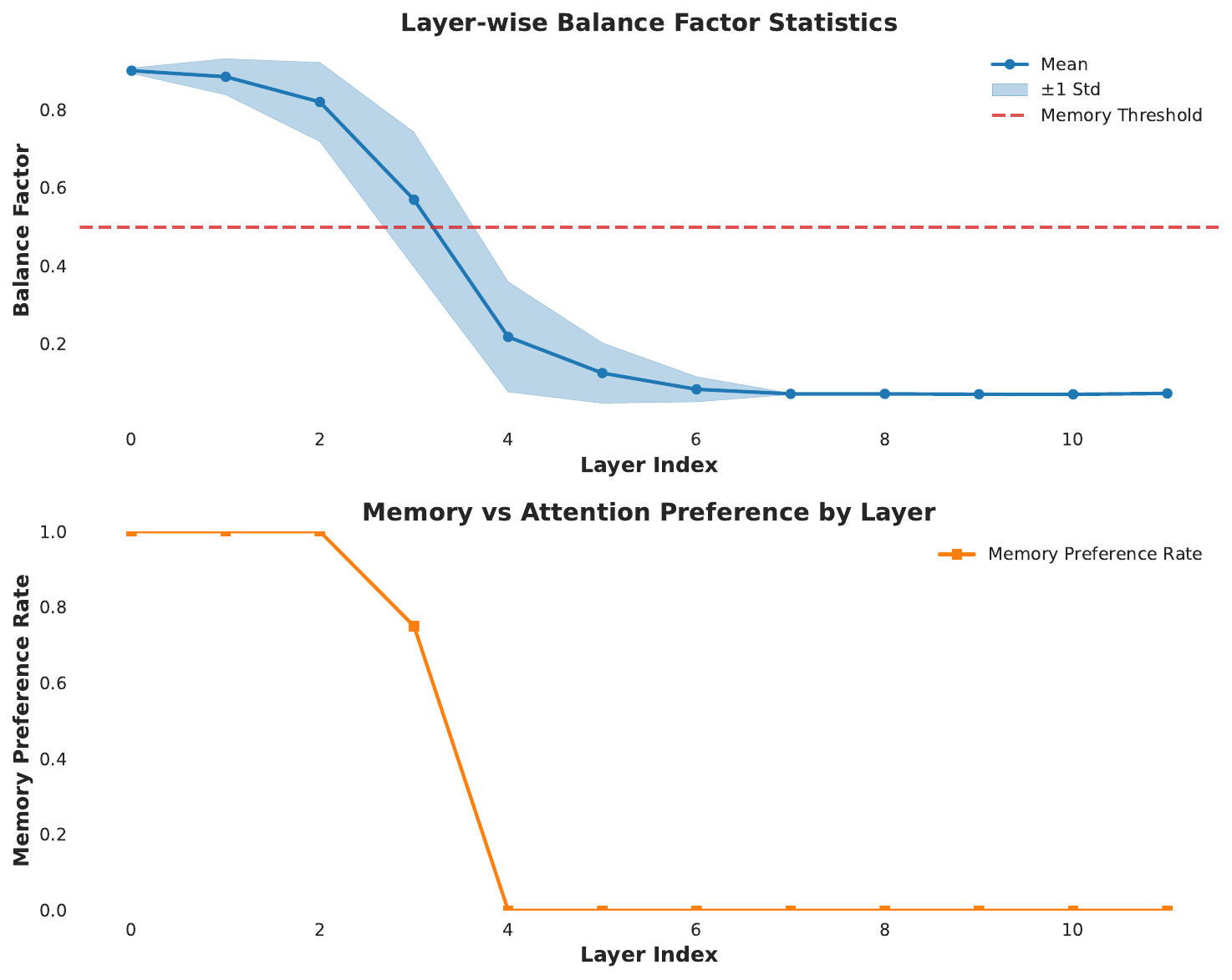}
    \caption{Layer-wise memory preference rate in Infini-attention. Lower layers show strong memory utilization, while higher layers show increased reliance on local attention.}
    \label{fig:layer_pref}
\end{figure}

To better understand how different layers utilize the memory mechanism, we analyze the layer-wise memory preference rates and the distribution of balance factors across the model architecture. Figure~\ref{fig:layer_pref} presents the memory preference rate across different layers. The analysis reveals distinct patterns in how different layers balance between local attention and memory retrieval mechanisms.

\subsubsection{Head-specific Balance Factor Analysis}
\label{sec:head_analysis}

\begin{figure}[hbt!]
    \centering
    \includegraphics[width=0.9\linewidth]{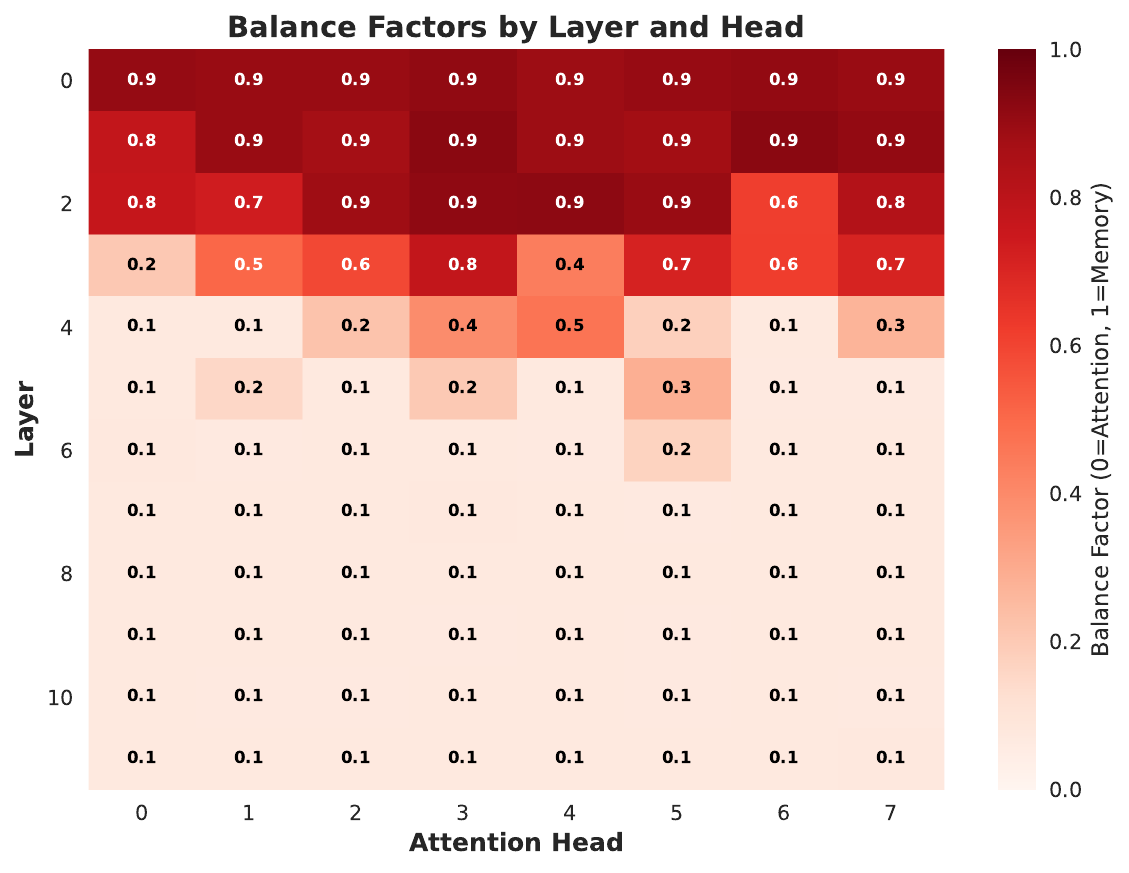}
    \caption{Heatmap of balance factors by layer and attention head. Darker shades of red indicate higher balance factors (more memory retrieval). The heatmap shows a clear pattern where lower layers contain more heads with high balance factors compared to higher layers.}
    \label{fig:bf_heatmap}
\end{figure}

Figure~\ref{fig:bf_heatmap} provides a comprehensive view of balance factors across all layers and heads, revealing the fine-grained attention patterns learned by the model.

\subsection{Additional Benchmark Results}
\label{sec:appendix_benchmarks}

\subsubsection{GLUE Benchmark Results}

The GLUE benchmark \citep{wang-etal-2018-glue} results show a mixed performance profile. The Infini-attention model excels on CoLA, MNLI, MRPC, QNLI, and WNLI tasks. Table~\ref{tab:glue_results} presents the detailed results.

\begin{table}[!htb]
\centering
\resizebox{\columnwidth}{!}{%
\begin{tabular}{lccccccccc}
\toprule
\textbf{Model} & \textbf{CoLA} & \textbf{MNLI} & \textbf{MNLI-mm} & \textbf{MRPC} & \textbf{QNLI} & \textbf{QQP} & \textbf{RTE} & \textbf{SST-2} & \textbf{WNLI} \\
\midrule
Baseline & 0.000 & 0.326 & 0.323 & 0.321 & 0.477 & \textbf{0.391} & 0.534 & \textbf{0.509} & 0.437 \\
& (0.000) & (0.005) & (0.005) & (0.023) & (0.007) & (0.002) & (0.030) & (0.017) & (0.059) \\
\midrule
Baseline (FT) & 0.000 & 0.325 & 0.324 & 0.355 & 0.474 & 0.385 & \textbf{0.538} & \textbf{0.509} & 0.437 \\
& (0.000) & (0.005) & (0.005) & (0.024) & (0.007) & (0.002) & (0.030) & (0.017) & (0.059) \\
\midrule
Infini-Attn & \textbf{0.080} & 0.327& \textbf{0.330} & \textbf{0.672} & \textbf{0.495} & 0.369 & 0.473 & 0.469 & \textbf{0.563} \\
& (0.027) & (0.005) & (0.005) & (0.023) & (0.007) & (0.002) & (0.030) & (0.017) & (0.059) \\
\midrule
Infini-Attn (FT) & 0.046 & \textbf{0.328} & 0.329& \textbf{0.672} & \textbf{0.495} & 0.367 & 0.473 & 0.471 & \textbf{0.563} \\
& (0.029) & (0.005) & (0.005) & (0.023) & (0.007) & (0.002) & (0.030) & (0.017) & (0.059) \\
\bottomrule
\end{tabular}%
}
\caption{Performance on GLUE benchmark. We report average of F1 score and accuracy for MRPC and QQP, Matthew's correlation coefficient for CoLA, and Accuracy for others. Results are shown with standard errors (±). Best results per task are highlighted in bold.}
\label{tab:glue_results}
\end{table}

\subsubsection{Long-Context Benchmark Results}

The benefits of Infini-attention are most pronounced on the long-context Scrolls benchmarks \citep{shaham-etal-2022-scrolls}. The fine-tuned Infini-attention model dramatically outperforms all other variants on both NarrativeQA and QMSum tasks, emphasizing its capability to work with long documents when appropriately fine-tuned. Table~\ref{tab:scrolls_combined} shows the results.

\begin{table}[!htb]
\centering
\resizebox{\columnwidth}{!}{%
\begin{tabular}{lcccc}
\toprule
& \textbf{NarrativeQA} & \multicolumn{3}{c}{\textbf{QMSum}} \\
\cmidrule(lr){2-2} \cmidrule(lr){3-5}
\textbf{Model} & \textbf{F1} & \textbf{ROUGE-1} & \textbf{ROUGE-2} & \textbf{ROUGE-L} \\
\midrule
Baseline & 0.000 & 0.033 & 0.008 & 0.033 \\
Baseline (Finetuned) & 0.019 & 0.853 & 0.006 & 0.853 \\
Infini-Attn & 0.068 & 0.049 & 0.000 & 0.049 \\
Infini-Attn (Finetuned) & \textbf{0.258} & \textbf{4.152} & \textbf{0.045} & \textbf{4.104} \\
\bottomrule
\end{tabular}%
}
\caption{Performance (\%) on the long-context Scrolls benchmark suite. Best results are highlighted in bold.}
\label{tab:scrolls_combined}
\end{table}

\end{document}